\documentclass{article}

\PassOptionsToPackage{numbers, sort}{natbib}
\usepackage[final]{bdl_2018}

\usepackage[utf8]{inputenc} 
\usepackage[T1]{fontenc}    
\usepackage{hyperref}       
\usepackage{cleveref}       
\usepackage{url}            
\usepackage{booktabs}       
\usepackage{amsfonts}       
\usepackage{nicefrac}       
\usepackage{microtype}      
\usepackage{macros}	
\usepackage{graphicx}
\usepackage{enumitem}

\title{Sequential Neural Methods \\ for Likelihood-free Inference}

\author{
  Conor Durkan \\
  University of Edinburgh \\
  \texttt{conor.durkan@ed.ac.uk} \\
  \And
  George Papamakarios \\
  University of Edinburgh \\
  \texttt{g.papamakarios@ed.ac.uk} \\
  \And
  Iain Murray \\
  University of Edinburgh \\
  \texttt{i.murray@ed.ac.uk}
}

\begin{document}

\maketitle

\begin{abstract}
Likelihood-free inference refers to inference when a likelihood function cannot be explicitly evaluated, which is often the case for models based on simulators. While much of the literature is concerned with sample-based `Approximate Bayesian Computation' methods, recent work suggests that approaches relying on deep neural conditional density estimators can obtain state-of-the-art results with fewer simulations. The neural approaches vary in how they choose which simulations to run and what they learn: an approximate posterior or a surrogate likelihood. This work provides some direct controlled comparisons between these choices.
\end{abstract}

\section{Introduction}
%

In this paper, we consider Sequential Neural Posterior Estimation \citep[SNPE]{papamakarios:2016:epsilon_free_inference, lueckmann2017flexible}, Sequential Neural Likelihood \citep[SNL]{papamakarios2018sequential}, and Active Sequential Neural Likelihood \citep[ASNL]{lueckmann2018likelihood}, three recent approaches to likelihood-free inference which leverage neural conditional density estimators to learn either a surrogate likelihood or approximate posterior directly. With each of the three models, we perform Bayesian inference over the parameters of the neural density estimators to prevent overfitting on small data sets, and in one case, investigate whether exploiting this uncertainty can lead to more efficient learning. Moreover, the modular nature of these methods allows us to perform direct comparison between them; though the implementation details of each have differed in the existing literature, here we use the same mixture density network architecture \citep[MDN]{bishop:1994:mdn} for each method, and train each MDN using stochastic variational inference (SVI)\@.

To our knowledge, this work provides the first head-to-head comparison of these sequential neural methods, and allows us to determine whether `uncertainty as a guiding principle' leads to more efficient or accurate inference for this problem. We find that, for our experiments, learning a surrogate likelihood is more efficient than learning the approximate posterior directly, and that using an active learning component based on Bayesian model uncertainty performs as well as the heuristic used by SNL, but provides no meaningful improvement.

\section{Problem \& Methods}
Formally, we are provided with a parametrized stochastic simulator $ p(\bfx \g \bftheta) $, which generates observations $ \bfx $ given parameters $ \bftheta $. It is possible to sample from $ p(\bfx \g \bftheta) $, but not evaluate this likelihood explicitly. Given a single observed data point $ \bfx_{0} $, we seek plausible settings of the simulator parameters $ \bftheta $ which generated that data. That is, we are interested in inferring the posterior distribution over parameters $ p(\bftheta \g \bfx_{0}) $. We are free to choose a suitable prior $ p(\bftheta) $. 

Each of the models under consideration proceeds in a sequential manner, alternating between the following two steps: 

\begin{enumerate}[leftmargin=0.5cm]
	\item \textbf{Generating data from the simulator}. A priori, and unlike standard machine learning tasks, we are provided with just a single observation $ \bfx_{0} $. We must choose parameter settings $ \bftheta $, and then rely on the simulator to generate pairs $ \curlybr{\roundbr{\bftheta^{(n)}, \bfx^{(n)}}}_{n=1}^{N} $ as training data.
        \item \textbf{Improving the posterior approximation}. Once training data has been gathered, we can use it to improve our posterior approximation. The choice of method will determine whether we model the forward relationship $ \bftheta \rightarrow \bfx $ (learn a surrogate likelihood, or emulator), or the inverse $ \bfx \rightarrow \bftheta $ (learn a recognition network that gives the posterior directly).
\end{enumerate}

\textbf{Sequential Neural Posterior Estimation (SNPE)}\@.
These methods fit a parametrized approximate posterior distribution $ q_{\bfphi}(\bftheta \g \bfx_{0}) $ to the true posterior $ p(\bftheta \g \bfx_{0}) $. In each round, the simulator is run using parameters sampled from the current approximate posterior \citep[SNPE-A]{papamakarios:2016:epsilon_free_inference}\@. Here we use the particular method by \citet{lueckmann2017flexible}, which we denote SNPE-B\@.

\textbf{Sequential Neural Likelihood (SNL)}\@.
In contrast to SNPE, SNL learns a parametrized surrogate likelihood $ q_{\bfphi}(\bfx_{0} \g \bftheta) \approx p(\bfx_{0} \g \bftheta) $. SNL also samples the parameters to simulate next from the current approximate posterior, which requires an additional inference step, such as Markov chain Monte Carlo using the surrogate likelihood. SNL also allows for easier use of other neural density estimators, so \citet{papamakarios2018sequential} use a state-of-the-art method MAF \citep{papamakarios:2017:maf}. Here we test whether SNL would still outperform SNPE-B if using the same neural architecture (in this case an MDN)\@.

\textbf{Active Sequential Neural Likelihood (ASNL)}\@.
Concurrently with SNL, \citet{lueckmann2018likelihood} proposed likelihood-free inference with emulator networks, a method which corresponds to SNL with an active learning component. We refer to this as active SNL, or ASNL\@. Rather than sampling from the approximate posterior each round to generate parameters $ \bftheta $, an acquisition function guides the choice of parameter at which to simulate next. The acquisition function exploits uncertainty in the approximate posterior, meaning we require a method to quantify uncertainty for the neural density estimator used as a surrogate likelihood. \citet{lueckmann2018likelihood} use the \textit{MaxVar} rule of \citet{jarvenpaa2018efficient}, given by $ \bftheta^{\star} = \argmax_{\bftheta} \V_{\bfphi} \squarebr{q_{\bfphi}(\bfx_{0} \g \bftheta) p(\bftheta)} $.


\section{Experiments}
We compare the three methods across three inference tasks (see \cref{experimental-details} for general details, \cref{asnl-details} for ASNL specific discussion, and \cref{additional-results} for additional results). 

\begin{figure}[!h]
	\centering
	\includegraphics[width=0.8\linewidth]{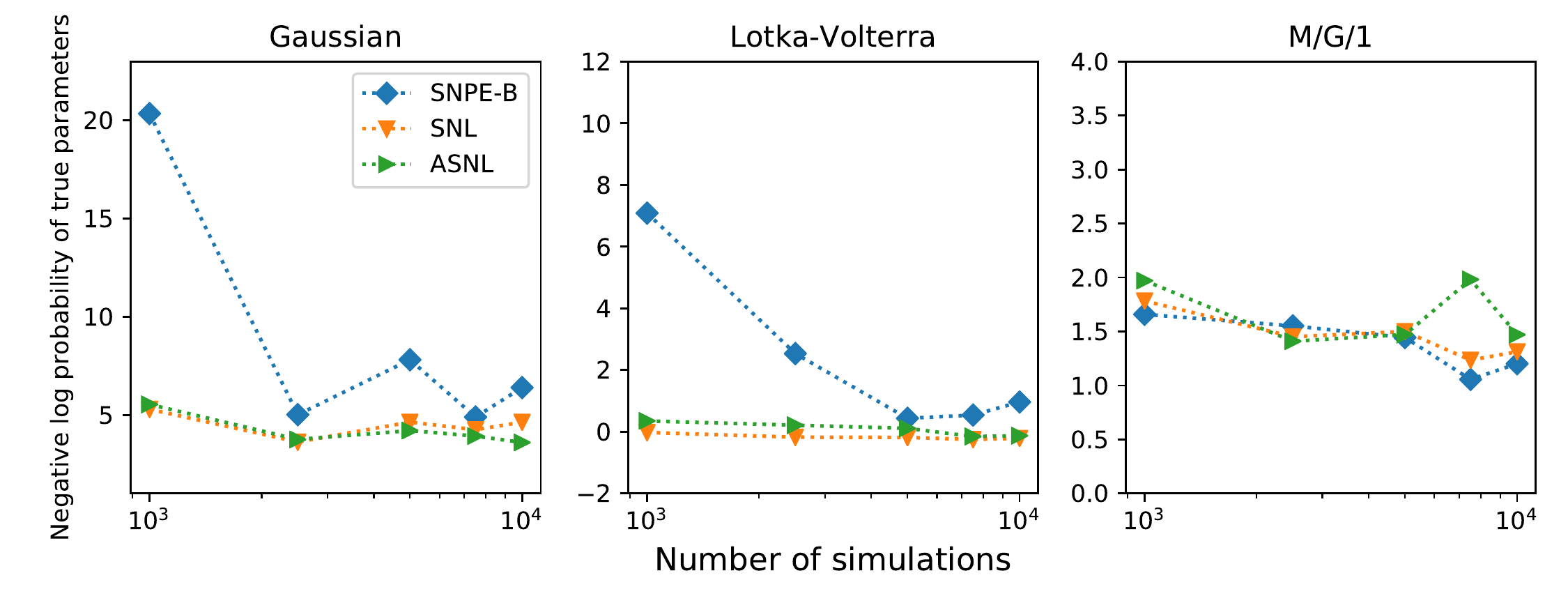}
	\caption{Comparison of true parameter log probability under the learned posteriors. Lower is better.}
	\label{log-prob}
\end{figure}

\begin{table}[!h]
	\caption{Wall-clock time (in hours) per experiment with $ 10^{4} $ simulations.}
	\label{sample-table}
	\centering
	\begin{tabular}{llll}
		\toprule
		     & Gaussian  & Lotka--Volterra & M/G/1 \\
		\midrule
		SNL & \textbf{\hphantom{0}7.48} & \textbf{\hphantom{0}8.27} & \textbf{\hphantom{0}7.83} \\
		\kern-1.3pt ASNL & 91.73 & 89.39 & 70.16 \\
		\bottomrule
	\end{tabular}
	\label{wall-clock-time}
\end{table}

\section{Discussion}
\textbf{Learning the likelihood rather than the posterior is a design choice}\@. Both SNL and ASNL achieve a higher log probability for the true parameters than SNPE-B on both the toy problem and Lotka--Volterra, while all three methods exhibit similar performance on M/G/1 (see \cref{log-prob}). This suggests learning the likelihood seems preferable to learning the posterior, all other model choices equal. However, the Gaussian task was chosen such that the likelihood was tractable, but the posterior was complex and multimodal. As such, it could be argued that the neural density estimator for SNPE-B in this task should have been made more powerful to compensate. 

Additionally, in both the Lotka--Volterra and M/G/1 models, we model summary statistics of the data, and not the observations themselves. Since summary statistics are generally much lower-dimensional than their associated observations, the burden on the neural density estimators modelling the likelihood is lessened. In contrast, direct posterior modelling with SNPE might converge faster than SNL when observations are relatively high-dimensional, and we need only learn a distribution over low-dimensional simulator parameters. 

On the other hand, the advance of general-purpose neural density estimators \citep{germain:2015:made, papamakarios:2017:maf}, in addition to task-specific models for images \citep{van2016pixelrnn} and audio \citep{van2016wavenet}, may mean that modelling the high-dimensional observations directly is more achievable than it once was. We might expect the final layers of such deep models to represent something akin to the summary statistics manually chosen for the data here.

\textbf{Quantifying uncertainty in neural networks is a difficult problem}\@. The success of any active learning component in these sequential neural methods is likely heavily dependent on the choice of uncertainty quantification. In this scenario, exploiting model uncertainty with active learning requires the ability to accurately and robustly quantify uncertainty over neural density estimators. Generally, this is a difficult problem, not just for neural density estimators, but also for the wider field of deep learning. Tools for uncertainty quantification in neural networks range from approximate Bayesian inference approaches like SVI \citep{hinton1993keeping, kingma:2015:variational_dropout, gal2016dropout}, expectation propagation \citep{hernandez2015probabilistic}, and Hamiltonian Monte Carlo \citep[HMC]{neal:2012:bnn}, to ensemble methods \citep{lakshminarayanan2017deepensembles}. Recent work \citep{depeweg2018uncertainty-decomposition, hafner2018uncertainty-noise-contrastive-priors, GalSmith2018Sufficient} has highlighted the concepts of \textit{aleatoric} (referring to stochasticity inherent to our data) and \textit{epistemic} uncertainty (traditional model uncertainty), discussing methods to disentangle these two confounding factors in deep learning. Despite this progress, robust, general-purpose uncertainty estimation for neural networks is still an open challenge for machine learning researchers. 

\textbf{Practical considerations influence our machine learning methods}\@. The active exploration of ASNL, though in theory more principled, does not seem to perform better than the sampling-based heuristic of SNL in practice. While being maximally efficient with simulations is preferable in principle, practical concerns may mean that wall-clock time is a more desirable metric. In this particular likelihood-free inference scenario, it could be argued that compute time taken by the ASNL acquisition function might be better allocated to additional simulation time for SNL, or a more complex neural density estimator (see \cref{wall-clock-time}).

In our experiments, the neural density estimators are small enough that a gold standard like HMC might be used to perform inference. However, we found HMC difficult to integrate with the models under consideration, being sensitive to initialization, and requiring manual tuning per round of training. We are also unaware of any work in which HMC is used to perform approximate inference over a neural density estimator, and it may be the case that HMC requires specific tuning for this scenario. In contrast, SVI performed well as a plug-in method, though the approximate posterior distribution it can represent is limited. Once more, there is a trade-off between less expressive but practical methods, and more complex, flexible methods which may be difficult for practitioners to tailor for their particular problems. 

\textbf{Conclusions}\@.
Broadly, we find:
\begin{itemize}[leftmargin=0.5cm]
	\item Even when restricting SNL to use the same SVI-MDN architecture as SNPE-B, SNL learns faster on the given tasks, and unlike SNPE-B, SNL allows for easy integration of other advanced neural density architectures. Future comparisons might focus on higher-dimensional data, which may lead to different conclusions.
	\item The heuristic exploration of SNL seems just as effective as the more principled exploration of ASNL\@. While these neural networks are useful for Bayesian inference tasks, whether principled uncertainty estimates are useful for guiding exploration remains to be seen.
\end{itemize}

\subsubsection*{Acknowledgments}
This work was supported in part by the EPSRC Centre for Doctoral Training in Data Science, funded by the UK Engineering and Physical Sciences Research Council (grant EP/L016427/1) and the University of Edinburgh. George Papamakarios was also supported by Microsoft Research through its PhD Scholarship Programme.


\bibliographystyle{apalike}

\appendix

\section{Experimental details}
\label{experimental-details}

\subsection{Bayesian MDN} 
The neural density estimator and its training procedure are fixed across methods for fair comparison. In all experiments, we use an MDN with a single hidden layer of 50 hyperbolic tangent units. The mixture distribution consists of 5 full-covariance Gaussian densities. We learn a diagonal Gaussian distribution over the weights and biases of the MDN using SVI and the local reparameterization trick \citep{kingma:2015:variational_dropout}, and use a standard normal prior. The density estimators are trained across 10 rounds using stochastic gradient descent with the Adam optimizer \citep{kingma2014adam}, with 1000 epochs per round.

\subsection{Posterior inference}
Both SNL and ASNL require an additional inference step to generate samples from the surrogate likelihood and prior. Following \citet{papamakarios2018sequential}, we used slice sampling, which performed robustly on each of our experiments.


In the interest of fair comparison, we used a kernel density estimate to approximate the log probability of the true parameters under the learned posterior, even though exact density evaluation of the posterior is available with SNPE-B\@. For each trained model, the kernel density estimate was based on 5000 samples from the posterior.

\subsection{Inference tasks}



\textbf{Gaussian toy problem}. The Gaussian problem outlined by \citet{papamakarios2018sequential} describes a toy simulator with a tractable (indeed, Gaussian) likelihood. Despite this simplification, the posterior is complex and multimodal. Such a model can be used to compare the advantages of learning the likelihood versus learning the approximate posterior directly.

\textbf{Lotka--Volterra}. The Lotka--Volterra model \citep{wilkinson2006lotkavolterra} describes the evolution of a predator-prey population over time. Four parameters $ \bftheta $ govern the birth and death rates of each population. The model is simulated using the Gillespie algorithm \citep{gillespie1977exact}, and observations $ \bfx $ are 9-dimensional summary statistics of the generated population time series. We follow the experimental setup of \citet{papamakarios:2016:epsilon_free_inference}. 

\textbf{M/G/1 Queue}. The M/G/1 queue model \citep{shestopaloff2014mg1} describes how a server processes a waiting queue of customers. Its three parameters $ \bftheta $ govern how long the server takes to respond to a customer and the frequency of customer arrivals. Observations $ \bfx $ are 5 equally-spaced quantiles of the distribution of inter-departure times. Again, we follow the experimental setup of \citet{papamakarios:2016:epsilon_free_inference}.

\section{ASNL Details}
\label{asnl-details}

\subsection{Modifications}
\citet{lueckmann2018likelihood} use an ensemble method \citep{lakshminarayanan2017deepensembles} to quantify uncertainty over the parameters of their density estimator, whereas we use SVI to maintain fair comparison to other methods. Their method also describes gathering a single data point at a time using the acquisition function, which we found not to provide a sufficient training signal for the neural density estimator. Instead, we allowed ASNL to generate as much data per round as given to SNPE-B and SNL, again for fair comparison.

\subsection{MaxVar implementation}
The \textit{MaxVar} acquisition function determines parameters $ \bftheta^{\star} $ at which to simulate by
\begin{align}
	\bftheta^{\star} = \argmax_{\bftheta} \V_{\bfphi} \squarebr{q_{\bfphi}(\bfx_{0} \g \bftheta) p(\bftheta)},
\end{align}
where the variance is taken with respect to the posterior distribution over neural density estimator parameters $ \bfphi $. This acquisition function focuses on areas of the parameter space where our estimate of the posterior varies the most. In each of our experiments, the prior on simulator parameters is box uniform (i.e.\ constant), so that the criterion can be equivalently written 
\begin{align}
\bftheta^{\star} = \argmax_{\bftheta} \V_{\bfphi} \squarebr{q_{\bfphi}(\bfx_{0} \g \bftheta)}.
\end{align}
We can approximate the function $ f(\bftheta) = \V_{\bfphi} \squarebr{q_{\bfphi}(\bfx_{0} \g \bftheta)} $ by sampling an ensemble of neural density estimators from the variational posterior $ q(\bfphi \g \curlyD) $, where $ \curlyD = \curlybr{\roundbr{\bftheta^{(n)}, \bfx^{(n)}}}_{n=1}^{N} $, and computing the sample variance of their density estimates for a particular value $ \bftheta $. However, neural density estimators generally yield log-densities, and in practice we found that the magnitude of these log-quantities was such that simply taking the exponent led to numerical issues. Fortunately, it is possible to sidestep this problem.

Computing probabilistic expressions in a numerically stable manner using log-quantities often involves the \textit{log-sum-exp} trick. Given log-quantities $ \{ a_{k} \}_{k=1}^{K} $, the log-sum-exp trick rewrites the naive computation as 
\begin{align}
	\log \sum_{k=1}^{K} \exp a_{k} = \max_{k} a_{k} + \log \sum_{k=1}^{K} \exp \squarebr{a_{k} - \max_{k^{\prime}} a_{k^{\prime}}}.
\end{align}
On the right hand side of this equation, the largest value in the summand is now 1, when $ a_{k} = \max_{k^{\prime}} a_{k^{\prime}} $. Any terms which now underflow are negligible when added to a value of at least 1, and so can safely be ignored.

We can also approximate a log-expectation of exponents with the log-sum-exp trick, since
\begin{align}
	\log \E \squarebr{\exp a} &\approx \log \squarebr{ \frac{1}{K} \sum_{k=1}^{K} \exp a_{k} } = \log \frac{1}{K} + \log \sum_{k=1}^{K} \exp a_{k}, \label{log-expec-exp}
\end{align} 
and this will prove useful in evaluating the MaxVar acquisition function.

Recall that we wish to find $ \bftheta $ which maximizes $ f(\bftheta) = \V_{\bfphi} \squarebr{q_{\bfphi}(\bfx_{0} \g \bftheta)} $. This is the same $ \bftheta $ which maximizes $ g(\bftheta) = \log \V_{\bfphi} \squarebr{q_{\bfphi}(\bfx_{0} \g \bftheta)} $, since the logarithm is monotonic. However, we only have sample values $ \log q_{\bfphi}(\bfx_{0} \g \bftheta) $, where $ \bfphi \sim q(\bfphi \g \curlyD) $. Phrasing the problem more generally for a non-negative random variable $ e^{a} $, we wish to approximate $ \log \V[e^{a}] $ in a numerically stable fashion, given log-samples $ a $ which may have large magnitude.

First rewrite 
\begin{align}
	\log \V[e^{a}] &= \log \E \squarebr{(e^{a} - \E[e^{a}])^{2}} \\
	&= \log \E \squarebr{\exp \roundbr{2 \log \vert e^{a} - \E[e^{a}] \vert}},
\end{align}
which is a log-expectation of exponents, and can be stably approximated by eq.~\eqref{log-expec-exp}. It remains to evaluate $ \log \vert e^{a} - \E[e^{a}] \vert $ in a stable manner, given that we have access to log-samples $ a $. To this end, write $ \E[e^{a}] = e^{b} $, where $ b = \log \E[e^{a}] $ can also be approximated by eq.~\eqref{log-expec-exp}, so that
\begin{align}
	\log \vertbr{ e^{a} - \E[e^{a}] } &= \log \vertbr{ e^{a} - e^{b} } \\
	&= \log \vertbr{ e^{a} \roundbr{ 1 - e^{b - a} } } \\
	&= a + \log \roundbr{1 - e^{b - a}} \hspace{0.2cm} \text{for } a > b.
\end{align}
Without loss of generality, we can assume $ a > b $, and the above expression can also leverage the numerically stable log-one-plus function. Thus, the final expression is given by
\begin{align}
	\log \V[e^{a}] = \log \E \squarebr{ \exp \roundbr{ 2 \roundbr{ a + \log \roundbr{1 - e^{b - a}} } } },
\end{align}
where $ b = \log \E[e^{a}] $, $ a = \log q_{\bfphi}(\bfx_{0} \g \bftheta) $, expectations are taken with respect to the distribution over neural density estimator parameters $ \bfphi $, and all log-expectation-exponent calculations can be approximated using eq.~\eqref{log-expec-exp}. We minimize the negative of this function using the L-BFGS-B optimizer \citep{zhu1997lbfgs} with constraints given by the box uniform priors on simulator parameters, and initializing each optimization run with a sample from the prior.

\section{Additional Results}
\label{additional-results}
\begin{figure}
	\centering
	\includegraphics[width=0.9\linewidth]{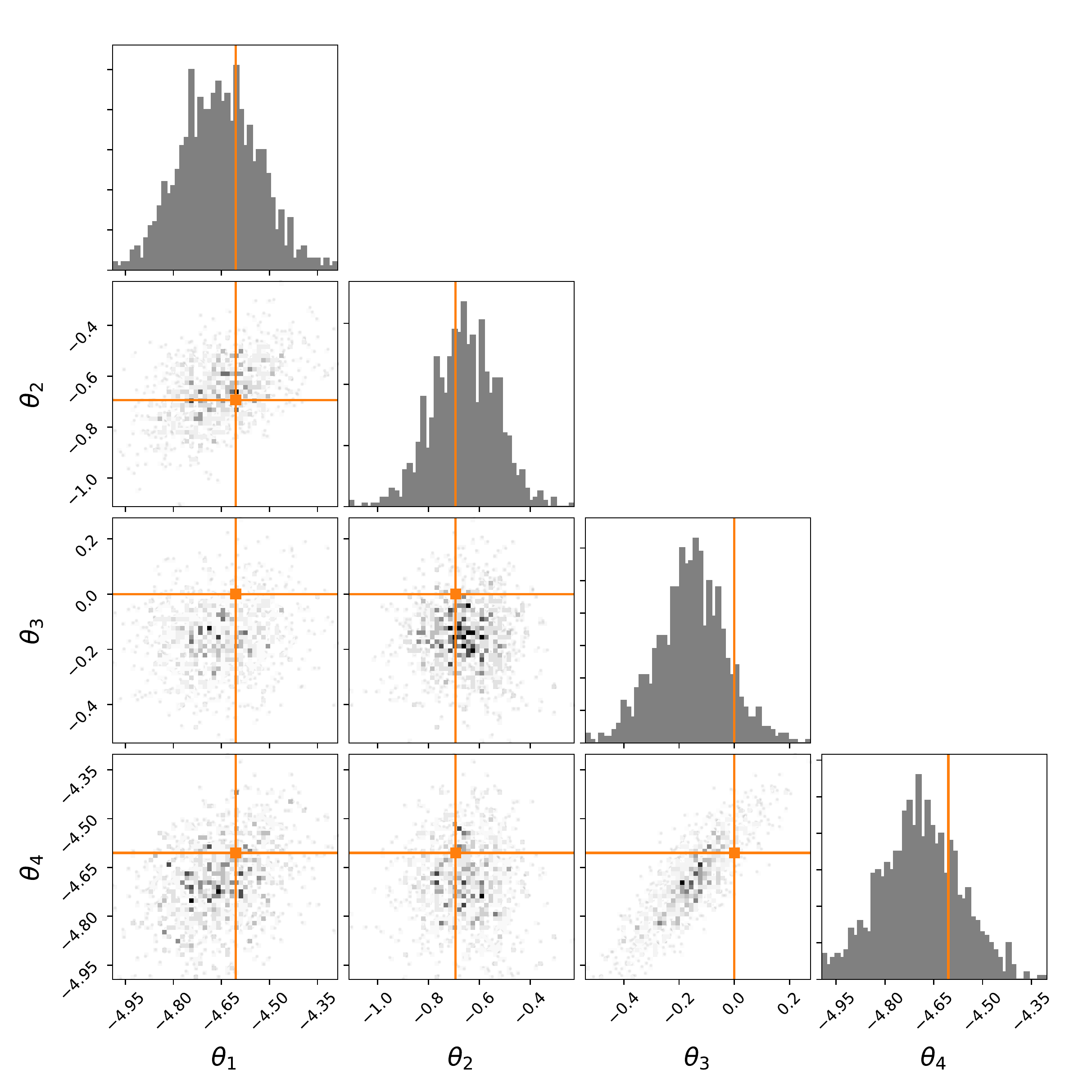}
	\caption{Simulator parameters generated using SNL before the final round of training for the Lotka--Volterra model. True parameters shown in orange. Since SNL generates parameters by performing MCMC on the current product of the approximate likelihood and prior, most of the gathered samples lie near the true parameter settings towards the end of training (compare to \cref{asnl-samples}).}
	\label{snl-samples}
\end{figure}

Though there seems to be no meaningful difference in the accuracy of the approximate posterior distributions learned by SNL and ASNL, it is interesting to compare the distributions of simulator parameters gathered by each of the methods towards the end of training. SNL, guided by the approximate posterior, should generate simulator parameters in a neighbourhood of the true parameters, if training has progressed suitably. From \cref{snl-samples}, it seems as if this is exactly the case.

\begin{figure}
	\centering
	\includegraphics[width=0.9\linewidth]{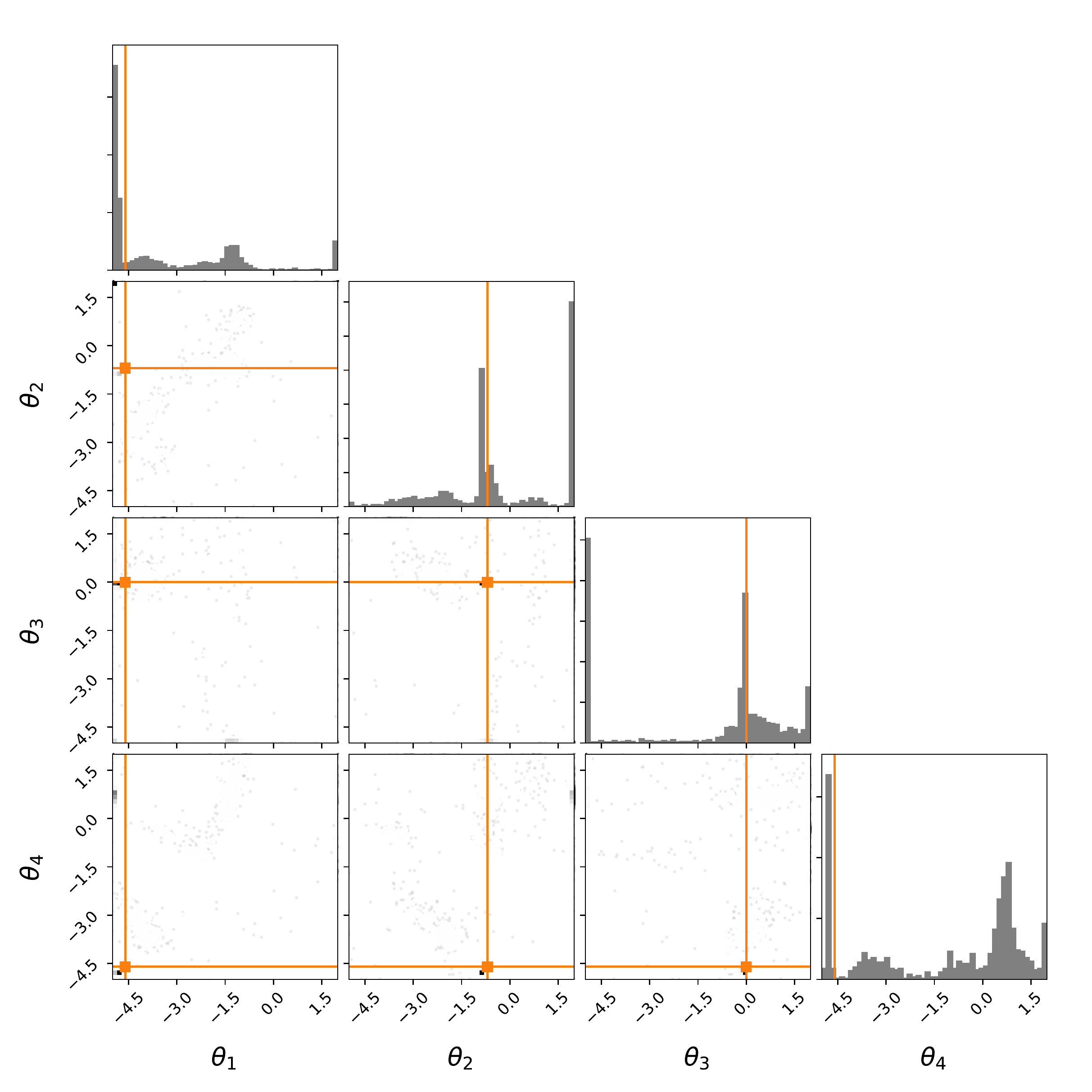}
	\caption{Simulator parameters generated using ASNL before the final round of training for the Lotka--Volterra model. True parameters shown in orange. ASNL generates simulator parameters using the MaxVar acquisition function, which focuses on regions of parameter space where our posterior approximation is most uncertain. Like SNL, many of the chosen parameters lie near the true values, but in contrast to SNL, ASNL seems also to be interested in the behaviour of the approximate posterior near the boundary defined by the box uniform prior (compare to \cref{snl-samples}).}
	\label{asnl-samples}
\end{figure}

In \cref{asnl-samples}, ASNL exhibits different behaviour. Note first the difference in range on each of the axes; SNL generates parameters very close the true parameters, while SNL generates parameters across the entire range of the box uniform prior $ [-5, 2]^{4} $. Though ASNL indeed focuses on neighbourhoods of the true parameters, it is also interested in other areas of parameter space. In particular, it generates many parameter settings near the boundary of the prior, and this is in line with results reported by \citet{lueckmann2018likelihood}. In some cases, it may be that this behaviour is desirable. While SNL is constrained to gather parameters where the likelihood is high, ASNL is free to explore the entire parameter space.

%

\end{document}